\pdfoutput=1
\documentclass[11pt]{article}

\usepackage{ACL2023}
\usepackage{times}
\usepackage{latexsym}

\usepackage[T1]{fontenc}
\usepackage[utf8]{inputenc}

\usepackage{microtype}

\usepackage{inconsolata}

\usepackage{graphicx} % 用于插入图片
\usepackage{subcaption} % 用于创建子图
\usepackage{booktabs} % 用于更好的表格线
\usepackage{adjustbox} % 用于调整表格大小
\usepackage{amsmath}
\usepackage{amssymb}
\usepackage{multirow}
\usepackage{colortbl}
\usepackage{booktabs}
\usepackage{bbding}
\usepackage{tabularx}
\usepackage{algorithm}
\usepackage{algpseudocode}
\definecolor{mygray}{gray}{0.9} % 阴影颜色

\usepackage{comment}

\newcommand{\cred}[1]{{\color{red}{#1}}}
\newcommand{\TODO}[1]{{\color{red} {\bf TODO: #1}}}

\definecolor{commentcolor}{RGB}{0,128,0}

\title{BitDistiller: Unleashing the Potential of Sub-4-Bit LLMs via Self-Distillation}
\author{
Dayou Du\textsuperscript{1}, Yijia Zhang\textsuperscript{2}, Shijie Cao\textsuperscript{3},
Jiaqi Guo\textsuperscript{3}, Ting Cao\textsuperscript{3}, 
Xiaowen Chu\textsuperscript{1}, Ningyi Xu\textsuperscript{2} \\
\textsuperscript{1}The Hong Kong University of Science and Technology (Guangzhou) \\
\textsuperscript{2}Shanghai Jiao Tong University \\
\textsuperscript{3}Microsoft Research Asia \\
ddu487@connect.hkust-gz.edu.cn, 
\{zhangyijia, xuningyi\}@sjtu.edu.cn, \\ 
\{shijiecao, jiaqiguo, ting.cao\}@microsoft.com, xwchu@ust.hk
}

\begin{document}
\maketitle
\begin{abstract}
\begin{comment}
deployment challenge 
-> quantization 

-> 4 bit as a standard, sub4bits, huge accuracy loss compared to full precision. 
( -> exsiting works PTQ + QAT, still a lot of potential for improvement)-> 

we propose xxx to unleash xxx, particularly in extreme low bit (3,2) settings    
\end{comment}

The upscaling of Large Language Models (LLMs) has yielded impressive advances in natural language processing, yet it also poses significant deployment challenges.
Weight quantization has emerged as a widely embraced solution to reduce memory and computational demands.  
This paper introduces BitDistiller, a framework that synergizes Quantization-Aware Training (QAT) with Knowledge Distillation (KD) to boost the performance of LLMs at ultra-low precisions (sub-4-bit).
Specifically, BitDistiller first incorporates a tailored asymmetric quantization and clipping technique to maximally preserve the fidelity of quantized weights, and then proposes a novel Confidence-Aware Kullback-Leibler Divergence (CAKLD) objective, which is employed in a self-distillation manner to enable faster convergence and superior model performance.
Empirical evaluations demonstrate that BitDistiller significantly surpasses existing methods in both 3-bit and 2-bit configurations on general language understanding and complex reasoning benchmarks.
Notably, BitDistiller is shown to be more cost-effective, demanding fewer data and training resources. The code is available at \url{https://github.com/DD-DuDa/BitDistiller}.

%BitDistiller tackles two fundamental challenges in low-bit QAT with KD: preserving the fidelity of quantized weights and effectively transferring knowledge in distillation. 

\end{abstract}

\section{Introduction}

%\TODO{for shijie: polish and citations}

Scaling up model sizes has been pivotal to the success of large language models (LLMs), yielding unprecedented performance across diverse natural language processing tasks~\cite{gpt-3,llama2,scaling_law}. 
However, such escalating model size poses significant challenges in deployment, particularly on resource-constrained devices, due to the substantial memory footprint and computational requirements.

Weight quantization has emerged as a popular strategy to enhance the efficiency and accessibility of LLMs by reducing model size with minimal performance loss~\cite{survey_quantization}.
In practice, 4-bit quantization has been widely adopted, offering a balance between a considerable compression ratio and the preservation of LLM capabilities~\cite{awq,gptq,llm-fp4}.

\begin{figure}[t] 
    \centering 
    \includegraphics[width=0.4\textwidth]{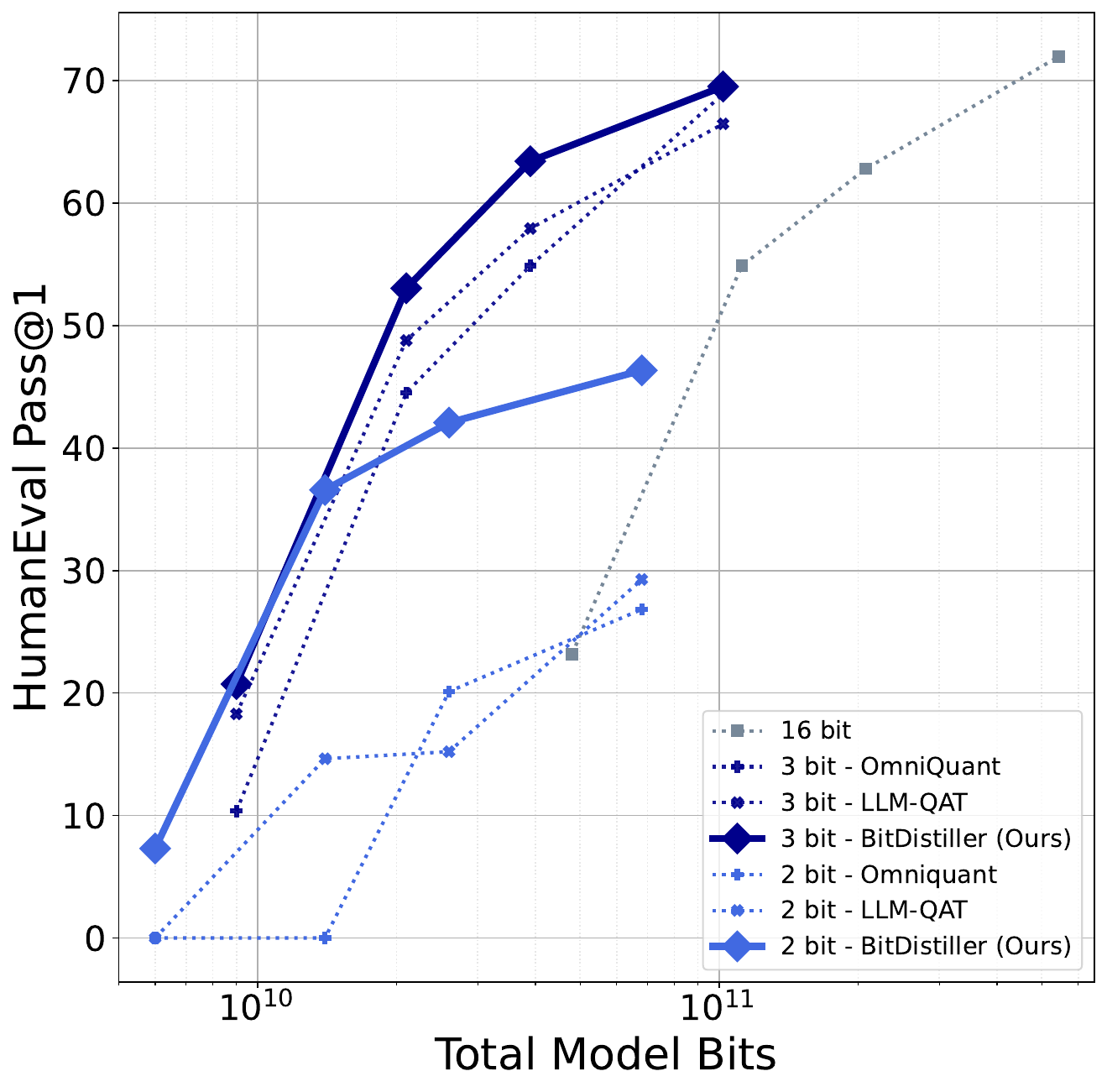} 
    \caption{Bit-Level scaling laws for code generation performance for 3B to 34B parameter coder models. BitDistiller outperforms existing QAT methods in both 3-bit and 2-bit settings. Details in Table~\ref{tab:reasoning-task}.} 
    \label{fig:scaling_law} 
\end{figure}

However, sub-4-bit quantization significantly degrades the fidelity of model weights, leading to deteriorated model performance, especially in smaller models or tasks requiring complex reasoning~\cite{scaling_law_quant}.
To address this, researchers have developed various Post-Training Quantization (PTQ) and Quantization-Aware Training (QAT) methods~\cite{quip,omniquant}.
PTQ, while appealing without retraining, struggles to preserve model performance at very low precisions.
In contrast, QAT incorporates quantization into the training loop, enabling dynamic adaptation to reduced precision and thus maintaining higher accuracy~\cite{llm-qat,tsld}.
Despite its early promise, two fundamental challenges are essential for achieving high model performance in extreme low-bit QAT: how to maximally preserve weight fidelity during quantization, and how to effectively learn low-bit representations during training.

%QAT for LLMs still faces significant challenges in extensive training and data requirements and remains largely uncharted, with vast potential for innovations and improvements.

In this work, we present BitDistiller, a novel framework that synergizes QAT with Knowledge Distillation (KD) to significantly boost the performance of sub-4-bit quantized LLMs. 
To minimize quantization error, BitDistiller employs a tailored asymmetric quantization and clipping strategy to maintain the capabilities of the full-precision model as much as possible, particularly at ultra-low-bit levels. For efficient and effective low-bit representation learning, BitDistiller leverages a simple yet effective self-distillation approach, wherein the full-precision model acts as its own teacher to refine the low-bit student model. Notably, BitDistiller innovates with a Confidence-Aware Kullback-Leibler divergence (CAKLD) objective that optimizes knowledge transferring efficacy, enabling faster convergence and enhanced model performance.

Our empirical evaluations, conducted on a diverse suite of general language understanding and complex reasoning tasks including mathematics and coding, demonstrate that BitDistiller significantly outperforms existing PTQ and QAT methods in the realm of sub-4-bit quantization. As illustrated in Figure~\ref{fig:scaling_law}, BitDistiller achieves the most favorable scaling law in both 3-bit and 2-bit configurations on the code reasoning benchmark. 
Moreover, BitDistiller is demonstrated to be more cost-effective, requiring less training data and fewer training resources, thereby marking a significant advancement toward deploying robust Large Language Models on resource-constrained devices.

\section{Background and Related Work}

%\TODO{The trend of extreme low bit, PB-LLM, BitNet}

\subsection{Weight Quantization for LLMs}
\paragraph{PTQ and QAT}
%Weight quantization for LLMs can be categorized into post-training quantization (PTQ) and quantization-aware training (QAT).
PTQ is directly applied to pre-trained models without additional training.
PTQ for LLMs typically employs techniques that either adjust quantization error~\cite{gptq,quip} or prioritize salient weights~\cite{spqr,awq,squeezellm}.
However, the lack of retraining with PTQ may cause notable decreases in model performance at extremely low precisions.
In contrast, QAT integrates quantization into the training phase, enabling the model to learn better representations for low-bit weights, as demonstrated by approaches like LLM-QAT \cite{llm-qat}, OmniQuant \cite{omniquant}, PB-LLM~\cite{PB-LLM}, and BitNet~\cite{bitnet}.
Despite improved model performance, QAT is still challenged by the need of extensive training and data, with significant potential for further optimization and enhancement.
In this work, we harness the synergy of QAT and KD to enhance the performance of quantized LLMs, especially at sub-4-bit settings. 

%\TODO{for shijie: one more sentence to discuss about the relationship of bitdistiller with previous work?}

\paragraph{Granularity and Format Optimizations}
%Quantization accuracy can be further improved by considering the granularity and format of quantization. 
Extensive research indicates that adopting finer-grained quantization approaches, such as group-wise quantization, can achieve higher accuracy compared to layer-wise or channel-wise methods~\cite{shen2020q, gptq}. 
Floating-point formats (FP8/FP4/NF4) have been demonstrated to deliver superior accuracy compared to integer formats (INT8/INT4) in LLM quantization~\cite{kuzmin2022fp8, scaling_law_quant, zhang2023integer}.
Notably, asymmetric quantization methods, particularly for floating-point formats, outperform their symmetric counterparts by better accommodating the distribution of model weights~\cite{afpq}. 
BitDistiller aligns with these insights, employing finer granularity and asymmetric techniques for quantization.

\subsection{Knowledge Distillation for LLMs} 
In the realm of LLMs, white-box knowledge distillation (KD) has become increasingly prevalent due to the accessible distribution of the teacher model, which facilitates the transmission of knowledge representations to the student model~\cite{hintonkd, surveyllmcompression}. Notably, MINILLM~\cite{minillm} utilizes the reverse KLD to ensure the accuracy and fidelity of language generation.
GKD~\cite{gkd} has explored alternative divergences called the generalized Jensen–Shannon divergence (JSD) and addressed the distribution mismatch by sampling outputs from the student model during training. 
%\TODO{jiaqi help proofread this paragraph}

To attain exceedingly high compression ratios, a promising method is to combine KD with model quantization, where KD can be effectively used to mitigate the accuracy decline of quantized models~\cite{ternarybert,kim2022understanding}. In cutting-edge research applying QAT-based KD for LLMs, TSLD~\cite{tsld} considers risks of overfitting and conducts logit distillation with ground truth loss. Similarly, LLM-QAT leverages randomly teacher-generated data for data-free distillation. In distinction from TSLD and LLM-QAT, we achieve better performance and cost-efficiency in the extremely low-bit quantization level.

\section{Methodology}

\begin{figure}[h] 
\centering 
\includegraphics[width=0.48\textwidth]{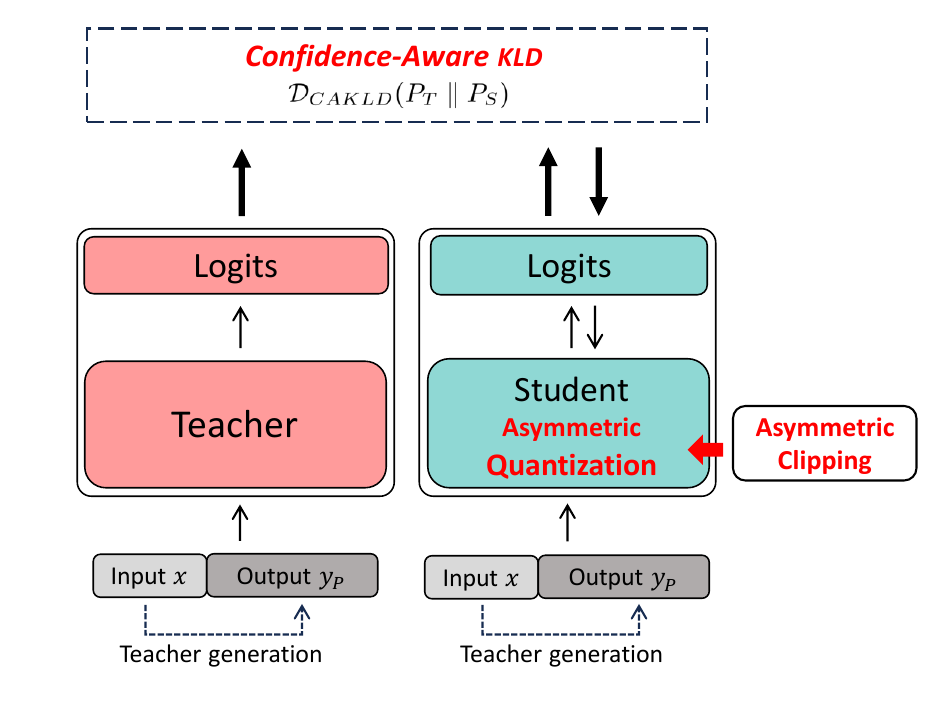} 
\caption{Depiction of the QAT-based KD framework of BitDistiller.} 
\label{fig:overview} 
\end{figure}

In this section, we introduce BitDistiller, a QAT with self-distillation framework for LLMs, as illustrated in Figure~\ref{fig:overview}.
To maximally preserve weight fidelity during quantization, we first present an asymmetric quantization and clipping method 
% to largely retains the full-precision counterpart's capabilities in extreme low-bit quantization
(see Section~\ref{subsec:method-asym}).
Second, to counteract the performance degradation caused by precision reduction, we adopt Knowledge Distillation and propose a novel Confidence-Aware KL divergence (CAKLD) objective, in which the full-precision model acts as a teacher and the low-precision one plays a student (see Section~\ref{subsec:method-kd}).

Algorithm~\ref{alg:overview} outlines the process of BitDistiller. Given the full-precision weight w, BitDistiller adopts the asymmetric clipping to alleviate outliers in w (Line~\ref{algline:clip}), prior to the training loop. Then, in each training step, BitDistiller forwards the model with the quantized weights ($w^t_Q$), computes the loss with the proposed CAKLD objective (Line~\ref{algline:quant}-\ref{algline:forward}), and updates the full-precision weights (Line~\ref{algline:backward}-\ref{algline:update})~\cite{ste}. When the training finishes, BitDistiller returns the final quantized weights.

% \subsection{Overview}
% We utilize full-precision pretrained or fine-tuned LLMs as teacher models, denoted by $P_T$, to guide the quantized student model $P_S^\theta$ equipped with trainable weights $\theta$. The corresponding quantized weights are denoted as $\hat{\theta} = Q_\theta(\theta)$, where $Q_\theta$ represents the asymmetric quantization function (refer to section \ref{subsec:method-asym}). Inspired by \cite{gkd, distillspec}, we employ a dataset comprising input-output sequence pairs $(X, Y)$, where $Y$ is the output generated by $P_T$. We initiate the training process by initializing $\theta$ with Asymmetric Clipping (as detailed in Eq.\ref{eq:asym_clip}) to more effectively distribute the quantized weights. Prior to training, we pre-calculate the coefficient $\beta$, essential for the Confidence-Aware KL Divergence (CAKLD), from a subset of $(X, Y)$. The training iterations are then conducted according to \cite{ste, ternarybert} with the keep of the full-precision weight, employing our proposed CAKLD as the guiding distillation loss. The comprehensive algorithm of BitDistiller is delineated in Algorithm \ref{alg:overview}.

% \TODO{for dayou: introduce the overview, what is teacher, what is student (low bit version with asym quant), what is the kd objective. explain algorithm 1. etc}
% refer: TernaryBERT: Distillation-aware Ultra-low Bit BERT

\begin{algorithm}
\caption{BitDistiller}
\begin{algorithmic}[1]
\item[\textbf{Input:}] Full-precision weight $w$, Dataset $\mathbb{D} = \{(x, y)\}$, Learning rate $\eta$, Training step $T$
\item [\textbf{Require:}] Clipping function $Clip$, Quantization function $Q$, Loss function $\mathcal{D}_\text{CAKLD}$
% Teacher model $P_T$ in Full-precision,
% \item[] Student Model $P_S^\theta$ with Asymmetric Quantization, \hfill \(\triangleright\)Eq.~\ref{eq:nf_2scale},Eq.~\ref{eq:int_scale_zero}
% \item[] Dataset $\mathbb{D} = \{(x, y)\}$,
% $(X, Y)$ contain input contexts $x$ and teacher output sequences $y$,
% \item[] Learning rate $\eta$.
% \item[\textbf{Initialize:}]
% Student weight $\hat{\theta}_1 = Clip(\theta)$
% Student weight $\theta$ with Asymmetric Clipping, \hfill \(\triangleright\)Eq.~\ref{eq:asym_clip}
% \item[] Coefficient $\beta$ from subset of $(X, Y)$.\hfill\(\triangleright\)Eq.~\ref{eq:beta}
\item[\textbf{Output:}] Low-precision weight $w^T_Q$ \\
$w^1 = Clip(w)$;\label{algline:clip}
\For{$t = 1$ \textbf{to} $T$}
    \State Sample a batch of data $\mathcal{B}$ from $\mathcal{D}$;
    % \State Quantize $\theta^t$ in student model to $\hat{\theta}^t$;
    \Statex \Comment{\textcolor{commentcolor}{Forward with quantized weight}}
    \State $w^t_{Q}$ = $Q(w^t)$;\label{algline:quant}
    \State Compute $\mathcal{D}_{\text{CAKLD}}(P^w \parallel P^{w^t_Q})$ on $\mathcal{B}$;\label{algline:forward}
    % $\mathcal{D}_{CAKL(\beta)}(P^T\|P_S^{\hat{\theta^t}})$; \hfill \% (\triangleright\)Eq.~\ref{eq:ca_kld}
    \Statex \Comment{\textcolor{commentcolor}{Backward on full-precision weight $w^t$}}
    \State Compute gradients $\frac{\partial \mathcal{D}_{\text{CAKLD}}}{\partial w^t}$;\label{algline:backward}
    \State $w^{t+1} = \text{Update}(w^t, \frac{\partial \mathcal{D}_{\text{CAKLD}}}{\partial w^t}, \eta)$\label{algline:update};
\EndFor\\
$w^T_Q$ = $Q(w^T)$;
\end{algorithmic}
\label{alg:overview}
\end{algorithm}

\subsection{Asymmetric Quantization and Clipping} \label{subsec:method-asym}
The adoption of finer granularities, or smaller group sizes, in weight quantization of LLMs inherently leads to asymmetrical distributions and the presence of outliers in weight groups. 
Proper management of asymmetry is crucial to maintaining model performance in low-bit PTQ regimes. 
Our investigation reveals that the effects of asymmetry are more prominent in extremely low-bit QAT, such as 3-bit and 2-bit configurations, necessitating tailored strategies to address these challenges.
Therefore, in BitDistiller, we adopt asymmetric quantization techniques coupled with asymmetric clipping strategies to enhance the representational fidelity of quantized weights and maximally preserve the capabilities of the full-precision model.

\paragraph{Asymmetric Quantization}

Previous studies have shown that floating-point formats (e.g., FP, NF) often outperform integer formats (INT) in LLM quantization~\cite{qlora,llm-fp4}. 
However, as the quantization level falls to 2-bit, we observed a notable decline in the effectiveness of FP/NF formats.
This advantage of FP/NF formats is attributed to their non-uniform nature, which can capture a wider range of values.
Such a non-uniform distribution aligns better with the natural distribution of weight tensors in LLMs.
In 2-bit cases, the limited representational capacity, offering only four distinct values, undermines the benefits of non-uniform distribution and impedes the efficient utilization of each numerical value.
In light of these findings, we employ NF formats for quantization above 2-bit, while opting for the INT format at the 2-bit level.

For NF formats (e.g., NF3), we adopt the AFPQ method~\cite{afpq} to enable asymmetric quantization, which establishes separate scales, $s_{pos}$ for positive weights $w_{pos}$ and $s_{neg}$ for negative weights $w_{neg}$, as shown in Equation~\ref{eq:nf_2scale}. 
For INT formats (e.g., INT2), we utilize conventional asymmetric methods with a single scale and a designated zero point, as detailed in Equation~\ref{eq:int_scale_zero}.

\begin{equation}
NF\text{-}Asym: Q(w) = 
\begin{cases} 
\lfloor \frac{w_{pos}}{s_{pos}} \rceil, & \text{if } w > 0 \\
\lfloor \frac{w_{neg}}{s_{neg}} \rceil, & \text{if } w \leq 0
\end{cases}
\label{eq:nf_2scale}
\end{equation}

\begin{equation}
INT\text{-}Asym: Q(w) = \lfloor \frac{w - z}{s} \rceil
\label{eq:int_scale_zero}
\end{equation}

\paragraph{Asymmetric Clipping} 

The strategy of clipping, which involves constraining the range of weight values, has been recognized for its contribution to maintaining high accuracy after quantization~\cite{sakr2022optimal,omniquant}. 
However, naive clipping methods often fall short in effectiveness, while advanced clipping techniques come at a high computational cost which is prohibitive for practical QAT use \cite{li2019fully, jung2019learning}.
To circumvent these limitations, we propose the use of asymmetric clipping solely during the initial phase, prior to the commencement of QAT. 
Asymmetric clipping at initialization provides a good starting point that significantly contributes to the final overall quantized model accuracy without incurring the prohibitive costs associated with iterative clipping optimization.

To enable asymmetric clipping for QAT initialization, given input features $X$ cached from a small calibration set, we conduct an automatic search for two optimal clipping values,
$\alpha$ and $\beta$, for each layer of the model. These values aim to minimize the output difference after quantization. Formally, the objective is to optimize the following:
\begin{equation}
    \begin{aligned}
        \alpha^{*},\beta^{*}&=\underset{\alpha,\beta}{\operatorname*{argmin}}||Q(w_{c}) X-w X|| \\&
        w_{c}=Clip(w,\alpha,\beta) \\&
        \begin{cases}
            \alpha\in[\min\_val,0)\\\beta\in(0,\max\_val]
        \end{cases}
    \end{aligned}
    \label{eq:asym_clip}
\end{equation}

\begin{figure}[h] 
\centering 
\includegraphics[width=0.45\textwidth]{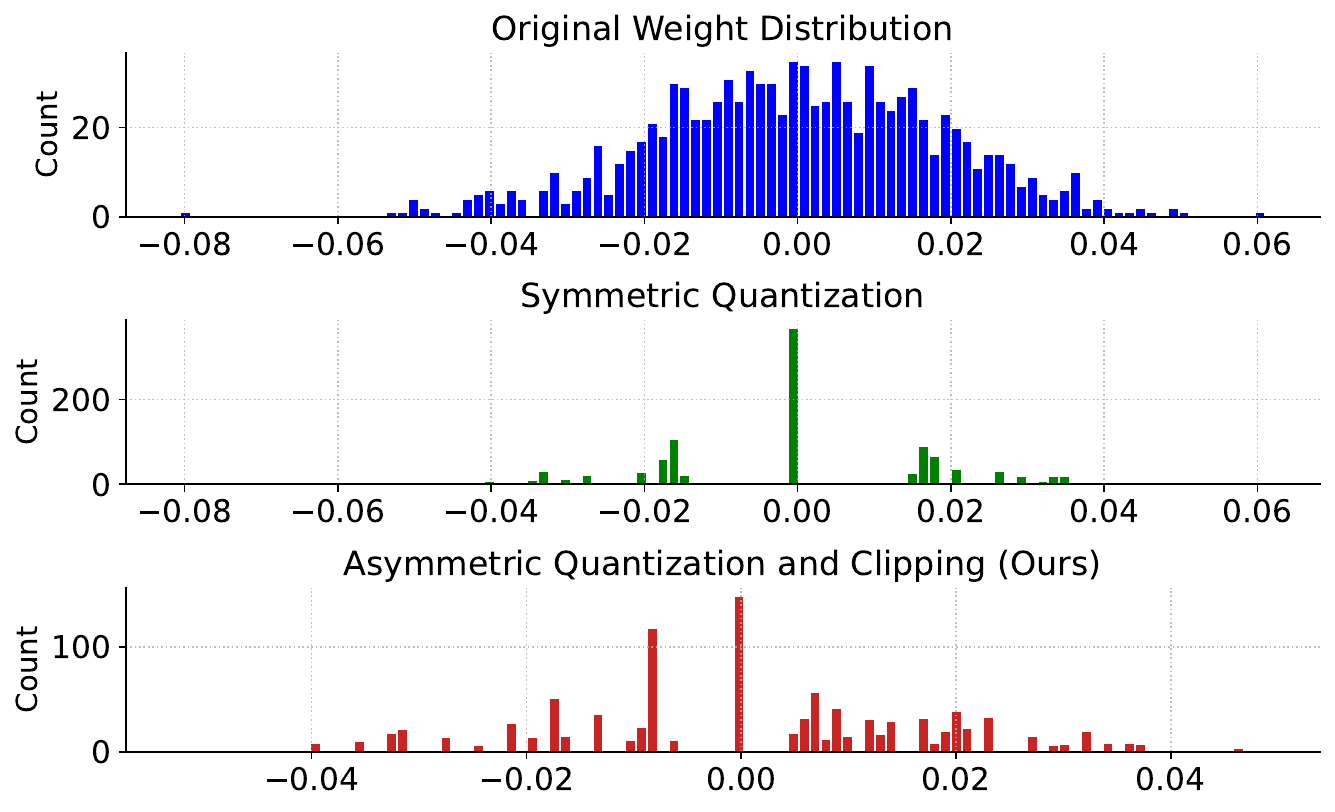} 
\caption{(Top) The original weight distribution of a single output channel in the final down projection layer of LLaMA-2-7B. (Middle\&Bottom) The weight distribution after symmetric quantization and asymmetric quantization and clipping, both using 3-bit quantization with the group size of 128.} 
\label{fig:asymmetric} 
\end{figure}

To demonstrate the efficacy of asymmetric quantization and clipping, we conduct a tensor-wise analysis. 
We selected a random weight tensor from the LLaMa-2-7B model and focused on a single output channel.
As illustrated in Figure~\ref{fig:asymmetric}, our approach to asymmetric quantization and clipping 
achieves higher fidelity preservation compared to symmetric quantization. 
A more detailed ablation study on the impact of asymmetric quantization and clipping on model performance is presented in Table~\ref{tab:ablation-asym} in Section~\ref{subsec:ablation}.

\begin{comment}

% Claim the outliers would hurt ...
While the scaling factor can be challenging to train \cite{llm-qat, omniquant}, we suggest that clipping the outlier value before QAT can contribute to improved performance. We automatically search for two clipping values, $\alpha$ and $\beta$, that minimize the output difference after quantization for a certain layer. Formally, our objective is to optimize the following objective:
\begin{equation}
    \begin{aligned}
        \alpha^{*},\beta^{*}&=\underset{\alpha,\beta}{\operatorname*{argmin}}||w_Q X-w X|| \\&
        w_{Q}=Q[clip(w,\alpha,\beta)] \\&
        \begin{cases}
            \alpha\in[\min\_val,0)\\\beta\in(0,\max\_val]
        \end{cases}
    \end{aligned}
    \label{eq:asym_clip}
\end{equation}

\end{comment}

\subsection{Self Distillation with CAKLD}
\label{subsec:method-kd}

To better counteract the performance degradation caused by precision reduction, we propose to adopt Knowledge Distillation (KD) in QAT, where the full-precision model acts as a teacher and its quantized variant plays a student:
\begin{equation}
    \mathcal{L} = \mathcal{D}(P_T \parallel P_S),
\end{equation}
where $\mathcal{D}$ is a divergence measure of two distributions. $P_T$ and $P_{S}$ denote the full-precision and quantized model, respectively.

The intuition for KD is two-fold.
First, learning the token-level probability distributions potentially helps the quantized model better imitate its full-precision counterpart~\cite{hintonkd}, thereby re-gaining the strong downstream performance.
Second, owing to the generative nature of LLM, it is easy to scale up the data size for QAT with the full-precision model.

% Due to the bit-reduction, the quantized LLMs would get relatively low capacity and worse performance compared with the full-precision counterpart.  Although Quantization Aware Training (QAT), which simulates lower precision during training to improve model robustness post-quantization, can help curb accuracy decline, relying solely on a fixed dataset with ground-truth output sequences for QAT presents several challenges. One significant issue is the limited variety and scale of data in a fixed dataset, which can restrict the model's ability to generalize, leading to overfitting. Therefore, for data-efficient and alleviate overfitting, we consider using knowledge distillation (KD) with QAT help compensate for accuracy degradation. More specifically, in contrast to larger-sized models, we use the original full-precision model as the teacher to achieve better alignment.

The divergence $\mathcal{D}$ chosen for distillation plays a crucial role.
\citet{gkd} find that the mode-seeking behavior advocated by the Reverse KL divergence (i.e., $\mathcal{D}_{KL}(P_S \parallel P_T$)) leads to better performance than Forward KL (i.e., $\mathcal{D}_{KL}(P_T \parallel P_S)$) on instruction tuning~\cite{chung2022scaling}, while Forward KL promotes mode-covering and is superior on general text generation tasks like summarization~\cite{narayan-etal-2018-dont}.
To provide a general receipt for QAT, we aim to seek a way to trade off the mode-seeking and mode-covering behaviors automatically, instead of manual selection according to some empirical understanding of downstream tasks.

\begin{figure}[t] 
\centering 
\includegraphics[width=0.25\textwidth]{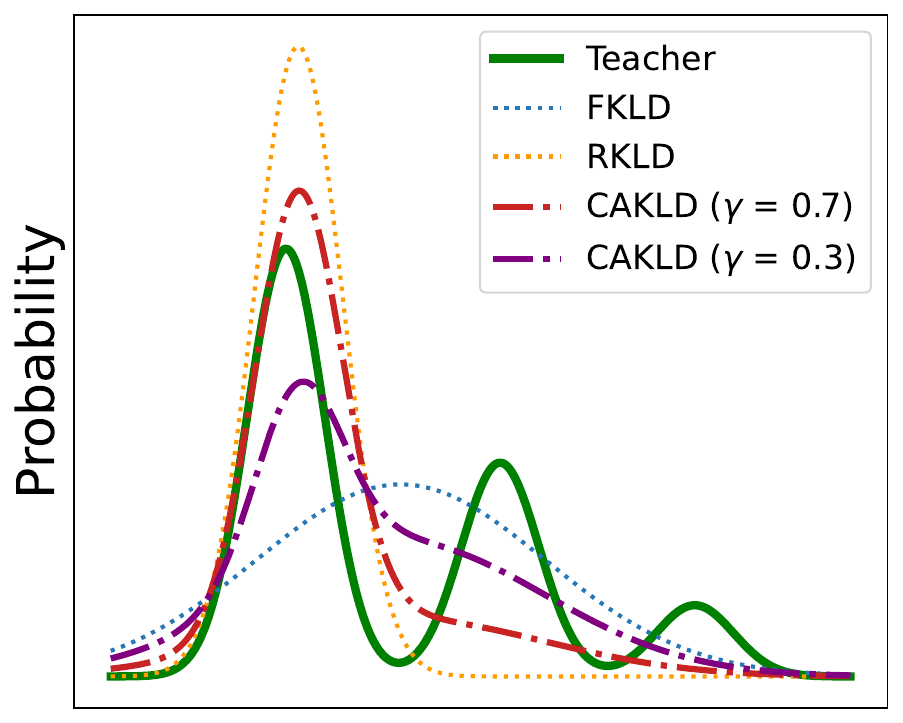} 
\caption{Comparison of Reverse KL, Forward KL and CAKLD, when a Gaussian distribution tries to fit a Gaussian mixture (Teacher).} 
\label{fig:cakld_demo} 
\end{figure}

To this end, we propose a novel \textbf{Confidence-Aware KL divergence}, shorted as CAKLD.
It blends the Reverse KL and Forward KL with a coefficient $\gamma$ estimated by the averaged token probability, so that 
the mode-seeking and mode-covering behaviors can be automatically traded off according to the full-precision model's confidence on the training data:
\begin{equation}
\small
\begin{aligned}
    \begin{split}
        \mathcal{D}_{CAKLD}(P_T\parallel P_S) &= \gamma\mathcal{D}_{KL}(P_S\parallel P_T)
        \\ &+(1-\gamma)\mathcal{D}_{KL} (P_T\parallel P_S)\\
    \end{split}\\
    \begin{split}
        \mathcal{D}_{KL}(P_T \parallel P_S) = & \mathbb{E}_{(x,y) \sim \mathbb{D}}[\frac{1}{|\{y\}|}\sum^{|\{y\}|}_{i=1} \\ &\mathbb{E}_{c\sim P_T(\cdot|x, y_{<i})}[\log{\frac{P_T(c|x,y_{<i})}{P_S(c|x,y_{<i})}}]]\\
    \end{split}\\
    \gamma = \mathbb{E}_{(x,y) \sim \mathbb{D}}[\frac{1}{|\{y\}|}\sum^{|\{y\}|}_{i=1}{P_T(y_i|x,y_{<i})}]
\end{aligned}
\label{eq:ca_kld}
\end{equation}
Intuitively, when the full-precision model is confident on the training data, CAKLD will prefer more on the mode-seeking behaviors.
Otherwise, CAKLD will advocate more on the mode-covering behaviors, as the full-precision model is not certain about the data and modeling its single mode is suboptimal.
Figure~\ref{fig:cakld_demo} visualizes the difference between Reverse KLD, Forward KLD and CAKLD when a Gaussian distribution tries to fit a Gaussian mixture.
It is clear that CAKLD manages to trade off mode-seeking and mode-covering behaviors with the coefficient. For a detailed performance comparison and in-depth analysis, please refer to Figure~\ref{fig:objectives} and Appendix~\ref{subsec:cakld_detail}

\section{Experiments}
\begin{table*}[t]
    \centering
    \resizebox{0.85\textwidth}{!}{
    %\begin{adjustbox}{margin=0.2em}
    \begin{tabular}{cc|c|c|cccc|c} 
    \toprule[\heavyrulewidth]
        \multicolumn{2}{c|}{\textbf{LLaMA-2-7B}} & PPL $\downarrow$ & MMLU (5s)  & PIQA & Hella. & Wino. & ARC-c & Avg \\ \midrule \midrule
        \multicolumn{2}{c|}{BF16} & 5.47 & 46.45 & 77.86 & 57.14 & 68.35 & 43.34 & 58.63 \\ \midrule
        ~ & RTN & 6.65 & 38.65 & 75.24 & 53.70 & 67.32 & 38.56 & 54.69 \\ 
        \multirow{2}{*}{3 Bits} & GPTQ & 6.38 & 39.57 & 75.46 & 51.68 & 67.16 & 38.39 & 54.45 \\ 
        \multirow{2}{*}{g128} & AWQ & 6.71 & 39.68 & 76.27 & 55.14 & 67.56 & 40.61 & 55.85 \\ 
        ~ & OmniQuant & 6.10 & 41.22 & 77.47 & 54.41 & 67.09 & 39.08 & 55.85 \\ 
        ~ & LLM-QAT & 6.02 & 41.32 & 77.26 & 54.74 & 68.35 & 40.61 & 56.46 \\ 
        ~ & \multicolumn{1}{>{\columncolor{mygray}}c|}{\textbf{BitDistiller (ours)}} & \multicolumn{1}{>{\columncolor{mygray}}c|}{\textbf{5.97}} & \multicolumn{1}{>{\columncolor{mygray}}c|}{\textbf{43.65}} & \multicolumn{1}{>{\columncolor{mygray}}c}{76.99} & \multicolumn{1}{>{\columncolor{mygray}}c}{55.38} & \multicolumn{1}{>{\columncolor{mygray}}c}{68.35} & \multicolumn{1}{>{\columncolor{mygray}}c|}{41.21} & \multicolumn{1}{>{\columncolor{mygray}}c}{\textbf{57.12}} \\ \midrule
        ~ & RTN & 3453 & 24.12 & 53.43 & 26.33 & 49.96 & 21.58 & 35.08 \\
        ~ & GPTQ & NaN & 23.12 & 49.51 & 25.04 & 49.57 & 22.69 & 33.99 \\
        2 Bits & AWQ & 2.2e5 & 25.38 & 52.39 & 25.70 & 50.12 & 21.33 & 34.98 \\
        g128 & OmniQuant & 12.84 & 25.42 & 58.92 & 29.20 & 50.83 & 19.45 & 36.76 \\
        ~ & LLM-QAT & 9.30 & 23.62 & 70.08 & 43.79 & 61.64 & 29.09 & 45.64 \\ 
        ~ & \multicolumn{1}{>{\columncolor{mygray}}c|}{\textbf{BitDistiller (ours)}} & \multicolumn{1}{>{\columncolor{mygray}}c|}{\textbf{8.08}} & \multicolumn{1}{>{\columncolor{mygray}}c|}{\textbf{29.25}} & \multicolumn{1}{>{\columncolor{mygray}}c}{73.61} & \multicolumn{1}{>{\columncolor{mygray}}c}{48.70} & \multicolumn{1}{>{\columncolor{mygray}}c}{61.09} & \multicolumn{1}{>{\columncolor{mygray}}c|}{33.27} & \multicolumn{1}{>{\columncolor{mygray}}c}{\textbf{49.18}} \\
    \bottomrule
    \end{tabular}
    }
    %\end{adjustbox}
    \caption{\textbf{General language task} results of BitDistiller versus established PTQ and QAT methods on LLaMA-2-7B Model. Our method achieves leading performance in both 3-bit and 2-bit quantization.}
    \label{tab:language_model}
\end{table*}

% \shijie{dayou, make sure we don't mis or mix use task name, metric name, and dataset name.}
We evaluate BitDistiller on the LLaMA-2 \cite{llama2} families and domain-specific LLMs with sub-4–bit quantization. We have set up comparative experiments to demonstrate the proficiency of our method against existing PTQ and QAT methods. Our findings illustrate that BitDistiller substantially enhances both the general language performance and the accuracy of reasoning tasks.

\subsection{Experimental Settings}
\paragraph{Tasks and Models} Following \cite{gptq, awq}, we benchmark LLaMA-2 \cite{llama2} on general language tasks, including language modeling tasks (WikiText-2 \cite{wikitext}), common sense QA benchmarks (PIQA \cite{piqa}, HellaSwag \cite{hellaswag}, WinoGrande \cite{winogrande}, ARC \cite{arc}) and in-context learning ability (MMLU \cite{mmlu}) under a few-shot setting. 
We also consider the complex reasoning tasks and evaluate various sizes of domain-specific LLMs, including WizardCoder \cite{wizardcoder} on LLM-Humaneval-Benchmarks \cite{humaneval} in the setting of greedy decode, and MetaMath \cite{metamath} on GSM8K \cite{gsm8k}. To evaluate the domain-specific LLMs of smaller sizes, we finetune OpenLLaMA-3B \cite{openllama} with domain-specific datasets. 

\paragraph{Baselines} PTQ baselines include vanilla round-to-nearest (RTN), GPTQ \cite{gptq}, AWQ \cite{awq}, Omniquant \cite{omniquant} and QuIP \cite{quip}. QAT baselines include LLM-QAT \cite{llm-qat} and TSLD \cite{tsld}. Detailed PTQ and QAT settings can be found in appendix \ref{subsec:ptq_qat_baselines}.

\paragraph{Quantization and Distillation} 
%We apply weight-only quantization to the linear layers in each decoder layer of the LLMs. 
We focus on 3-bit/2-bit group-wise quantization, with a group size of 128 (represented as 'g') as the default setting except for the 3B models with a group size of 64 because of the dimension constraint.
Following \cite{llm-qat, tsld}, we utilize logits distillation. Prior to QAT, the coefficient $\gamma$, key for CAKLD, is pre-calculated from a subset of $\mathbb{D}$. The implementation details and example analysis of CAKLD are available in Appendix \ref{subsec:cakld_detail}.

% \cred{As for improving performance/model size tradeoff \cite{scaling_law_quant}, we used channel-wise for 4bit quantization and group sizes of 128, 64 for 3bit and 2bit, respectively, except otherwise specified. }
% \shijie{modify this part. should be something like: Our evaluation centers on 3-bit(NF3) and 2-bit(INT2). we use group size of 128 for both.}

\paragraph{Training Datasets}
We use the instruction-tuning data from Alpaca \cite{alpaca} and the training set of WikiText-2 for general language tasks. 
For code understanding and generation, we use Evol-Instruct-Code \cite{evol_teacher}. For math reasoning we use MetaMathQA \cite{metamath}.

Given the instruction prompt $x$, sequence $s = \{x, y\}$ where output $y\sim p(\cdot|x)$ have three different choices:  Ground Truth $y_g$, Student-generated Output $y_q$ and Teacher-generated Output $y_p$. As suggested by \cite{gkd, distillspec}, we opt to generate the Teacher-generated Output $y_p$ using sampling with a temperature of 0.7 \cite{gsm8k_rft}. We conduct experiments in Section \ref{sec:data_generation} for an ablation study on the choices of output $y$. (See Appendix \ref{sec:training_data} for more details of training datasets composition).

\paragraph{Training Implementation}
We leverage DeepSpeed \cite{deepspeed} and HuggingFace repository~\cite{wolf-etal-2020-transformers} to devise a QAT-based KD framework enabling the distillation of models up to 34B. The model optimization is facilitated through the AdamW optimizer \cite{adamw}, applied with zero weight decay. We initialize the constant learning rate to 8e-6 and set the sequence length to 1024 for the code-related task and 512 for others.

\subsection{Evaluation on Language Modeling Tasks}

Table \ref{tab:language_model} presents a comparative analysis of BitDistiller's performance against previous PTQ and QAT methods on general language tasks. BitDistiller surpasses competing methods in terms of WikiText-2 perplexity and MMLU (5-shot) accuracy. Furthermore, BitDistiller demonstrates consistent performance across various QA benchmarks. Notably, in 2-bit weight quantization, BitDistiller substantially increases the average accuracy by +3.54\% over LLM-QAT \cite{llm-qat} and by +12.43\% compared to the leading PTQ method \cite{omniquant}. Similar results on LLaMA-2-13B can be found in Table \ref{tab:language_model_13b} in the Appendix \ref{subsec:appendix_13b}.

\subsection{Evaluation on Reasoning Tasks}
\begin{table*}[h]
    \centering
    \resizebox{0.85\textwidth}{!}{
    \begin{tabular}{cc|cccc|ccc}
    \toprule
        \multicolumn{2}{c|}{\multirow{2}{*}{\textbf{Domain-specific LLMs}}} & \multicolumn{4}{c|}{HumanEval @WizardCoder} & \multicolumn{3}{c}{GSM8K @MetaMath} \\ 
        ~ & ~  & 3B & 7B & 13B & 34B & 3B & 7B & 13B \\ \midrule \midrule
        \multicolumn{2}{c|}{BF16} & 23.17  & 54.88 & 62.80 & 71.95 & 36.40 & 66.41 & 72.30 \\ \midrule  
        ~  & RTN  & 4.27 & 34.15 & 50.00 & 33.54 & 17.50 & 59.30 & 68.51 \\ 
        ~  & GPTQ  & 4.30 & 46.34 & 55.48 & 63.41 & 6.72 & 62.11 & 68.75 \\ 
        3 Bits & AWQ  & 16.46 & 45.73 & 53.04 & 67.07 & 21.87 & 62.34 & 68.67 \\ 
        g128 & OmniQuant  & 10.36 & 44.51 & 54.88 & 68.90 & 23.67 & 61.70 & 68.28 \\ 
        ~ & LLM-QAT & 18.29 & 48.78 & 57.92 & 66.46 & 26.25 & 60.78 & 66.62 \\ 
        ~ & \multicolumn{1}{>{\columncolor{mygray}}c|}{\textbf{BitDistiller (ours)}}  & \multicolumn{1}{>{\columncolor{mygray}}c}{\textbf{20.73}} & 
        \multicolumn{1}{>{\columncolor{mygray}}c}{\textbf{53.66}} & 
        \multicolumn{1}{>{\columncolor{mygray}}c}{\textbf{63.41}} & 
        \multicolumn{1}{>{\columncolor{mygray}}c|}{\textbf{69.51}} & 
        \multicolumn{1}{>{\columncolor{mygray}}c}{\textbf{32.50}} & 
        \multicolumn{1}{>{\columncolor{mygray}}c}{\textbf{64.38}} & 
        \multicolumn{1}{>{\columncolor{mygray}}c}{\textbf{69.69}} \\ \midrule
        ~ & RTN & 0.0 & 0.0 & 0.0 & 0.61 & 0.0 & 0.0 & 7.89 \\ 
        ~ & GPTQ & 0.0 & 0.0 & 1.83 & 3.65 & 0.0 & 0.0 & 11.43 \\ 
        2 Bits & AWQ & 0.0 & 0.0 & 0.0 & 0.0 & 0.0 & 0.0 & 7.89 \\ 
        g128 & OmniQuant & 0.0 & 0.0 & 20.12 & 26.83 & 0.0 & 0.0 & 9.45 \\ 
        ~ & LLM-QAT & 0.0 & 14.63 & 15.21 & 29.27 & 6.56 & 23.13 & 36.64 \\ 
        ~ & \multicolumn{1}{>{\columncolor{mygray}}c|}{\textbf{BitDistiller (ours)}} & \multicolumn{1}{>{\columncolor{mygray}}c}{\textbf{7.31}} & 
        \multicolumn{1}{>{\columncolor{mygray}}c}{\textbf{36.59}} & 
        \multicolumn{1}{>{\columncolor{mygray}}c}{\textbf{42.07}} & 
        \multicolumn{1}{>{\columncolor{mygray}}c|}{\textbf{46.34}} & 
        \multicolumn{1}{>{\columncolor{mygray}}c}{\textbf{16.09}} & 
        \multicolumn{1}{>{\columncolor{mygray}}c}{\textbf{51.02}} & 
        \multicolumn{1}{>{\columncolor{mygray}}c}{\textbf{61.33}} \\ \bottomrule

    \end{tabular}
    }
    \caption{\textbf{Reasoning task} results of BitDistiller versus established PTQ and QAT methods on domain-specific LLMs. Our method achieves leading performance in both 3-bit and 2-bit quantization.}
    \label{tab:reasoning-task} 
\end{table*}
%\cred{should ptq and qat methods be marked in this table?}

Table \ref{tab:reasoning-task} demonstrates the superior performance of BitDistiller on reasoning-based benchmarks, including HumanEval and GSM8K, across a range of domain-specific language model families.
% Notably, BitDistiller consistently outshines competing quantization methods, delivering robust results even when models are constrained to a stringent 2-bit precision. 
BitDistiller achieves improvements over other methods in both 3-bit and 2-bit quantization.
Especially in 2-bit quantization, while other methods exhibit significant performance drops, BitDistiller maintains a commendable level of accuracy. 
Detailedly, our method outperforms LLM-QAT by a remarkable margin of 24.69\%, achieving an accuracy of 61.33\% on complex mathematical reasoning tasks. 
These outcomes bolster the potential for implementing ultra-low-precision inference deployment in practical reasoning tasks without substantially compromising performance.

\subsection{Ablation Studies} \label{subsec:ablation}

\paragraph{Asymmetric Quantization and Clipping}
\begin{table}[h]
    \centering
    \resizebox{\columnwidth}{!}{
    \begin{tabular}{cc|c|c}
    \toprule
        \multicolumn{2}{c|}{\multirow{2}{*}{\textbf{LLaMA-2-7B}}} & PPL $\downarrow$& MMLU (5s) $\uparrow$ \\ 
        ~ & ~ & (start $\mapsto$ end ) & (start $\mapsto$ end ) \\ \midrule \midrule
        \multirow{2}{*}{3 Bits}& NF-Sym & 6.45 $\mapsto$ 6.10 & 38.28 $\mapsto$ 39.27 \\ %\midrule % NF
        \multirow{2}{*}{g128}& $\rightarrow$ NF-Asym & 6.30 $\mapsto$ 6.01 & 41.53 $\mapsto$ 42.61 \\ %\midrule
        ~& $+$ Clip-Asym & 6.08 $\mapsto$ \textbf{5.97} & 42.90 $\mapsto$ \textbf{43.65} \\ \midrule
        \multirow{2}{*}{2 Bits} & INT-Sym & 2.4e5 $\mapsto$ 2.5e5 & 24.95 $\mapsto$ 26.03 \\ 
        \multirow{2}{*}{g128} & $\rightarrow$ INT-Asym & 3.4e2 $\mapsto$ 16.94 & 24.12 $\mapsto$ 24.82 \\ 
        ~ & $+$ Clip-Asym & 17.98 $\mapsto$ \textbf{8.08} & 26.75 $\mapsto$ \textbf{29.25} \\ \bottomrule
    \end{tabular}
    }
    \caption{Ablation study of asymmetric quantization and clipping on WikiText2 perplexity and MMLU (5-shot). The "start $\mapsto$ end" notation denotes the metric values before and after training.}
    \label{tab:ablation-asym} 
\end{table}

In this ablation study, we evaluate the efficacy of quantization strategies on the LLaMA-2-7B model. Our approach examines the impact of asymmetric quantization and clipping techniques within QAT. We specifically assess the 3-bit and 2-bit quantization levels, reporting our findings in terms of Perplexity (PPL) and MMLU (5-shot).

As demonstrated in Table \ref{tab:ablation-asym}, asymmetric quantization significantly enhances model performance. Notably, under a 2-bit configuration, PPL can be reduced from 3.4e2 to 16.94 in post-training. Furthermore, the application of asymmetric clipping during initialization yields additional performance gains upon training completion. 
See Appendix~ \ref{subsec:awq_init} for integration with other PTQ methods.

\paragraph{Data Generation} \label{sec:data_generation}

\begin{figure}[h]  
    \centering  
    \begin{subfigure}[t]{0.23\textwidth}  
        \includegraphics[width=\textwidth]{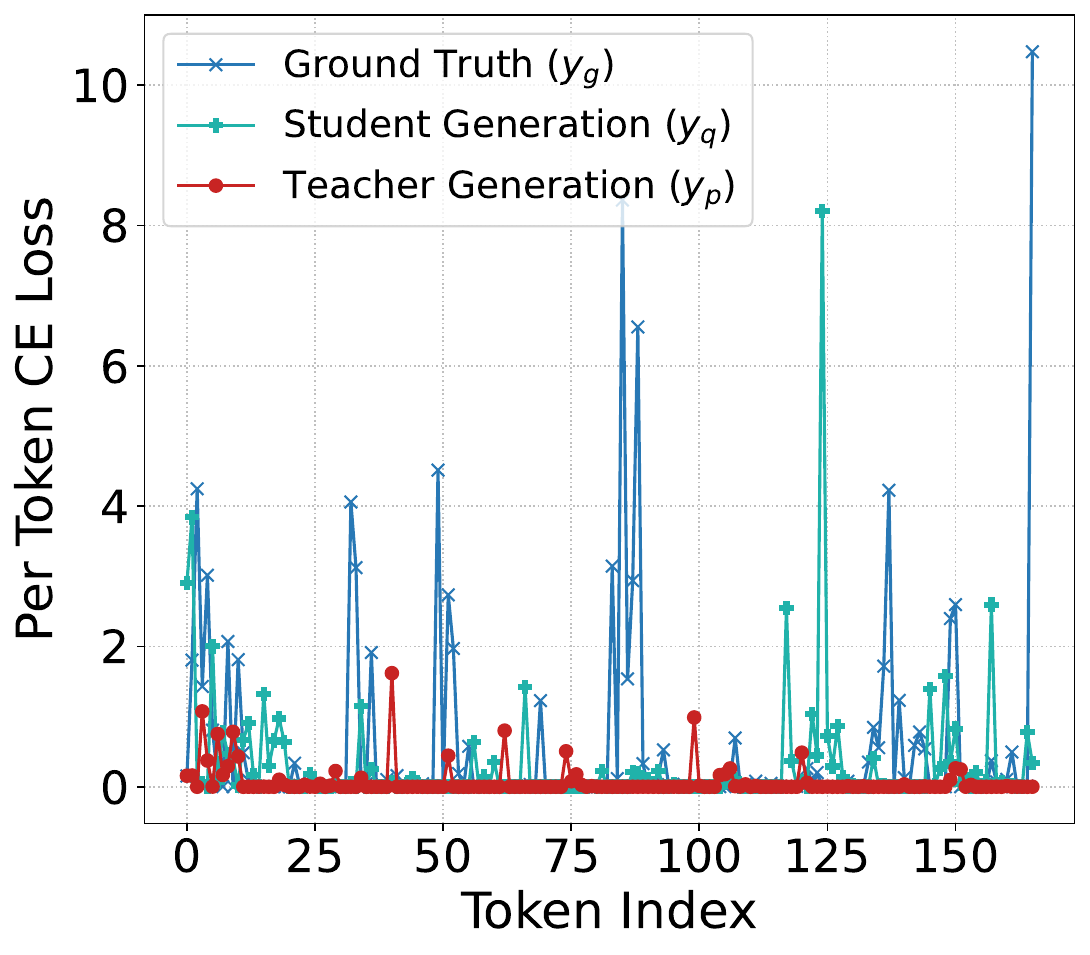} 
        \caption{}
        \label{fig:data_celoss}
    \end{subfigure}  
    \hfill
    \begin{subfigure}[t]{0.23\textwidth}  
        \includegraphics[width=\textwidth]{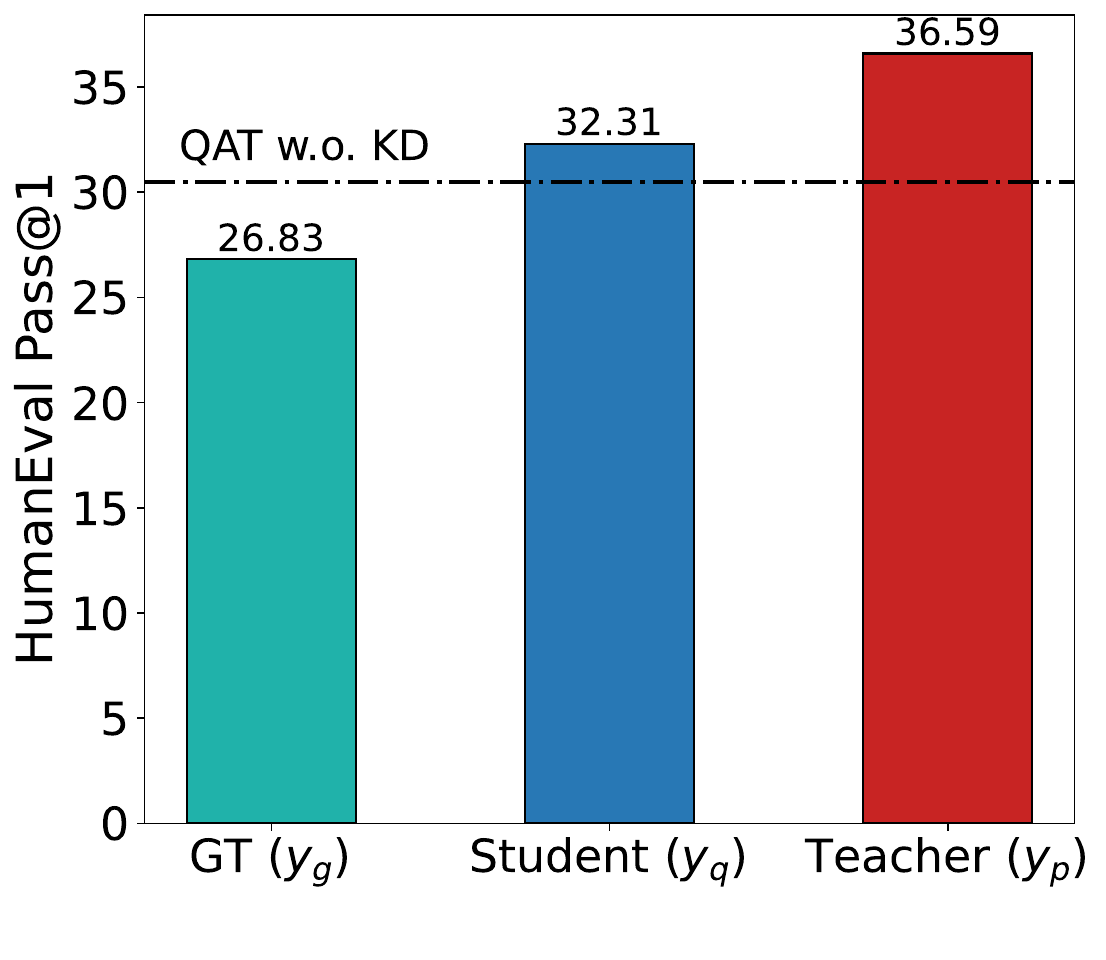}
        \caption{}
        \label{fig:data_generation}
    \end{subfigure} 
    \caption{Comparative analysis of using various data generation methods on WizardCoder-7B. (a) shows the per-token cross-entropy loss. (b) presents the HumanEval Pass@1. (‘QAT w.o. KD' indicates the baseline where only the ground truth dataset is used for supervised fine-tuning, without knowledge distillation.)} 
\end{figure} 

In our analysis, we meticulously evaluated the logit information of the teacher model by computing the cross-entropy loss (CELoss) for various outputs $y$. Figure \ref{fig:data_celoss} illustrates that the data generated by the teacher model $y_p$ exhibits low CELoss, indicative of a high-confidence logit distribution, which in turn facilitates better convergence with our proposed CAKLD. The comparative performance results depicted in Figure \ref{fig:data_generation} reveal that the use of teacher-generated data in conjunction with CAKLD yields superior outcomes when compared to employing either a fixed dataset or student-generated data $y_q$.

\paragraph{Distillation Objectives}

\begin{figure}[h] 
\centering 
\includegraphics[width=0.3\textwidth]{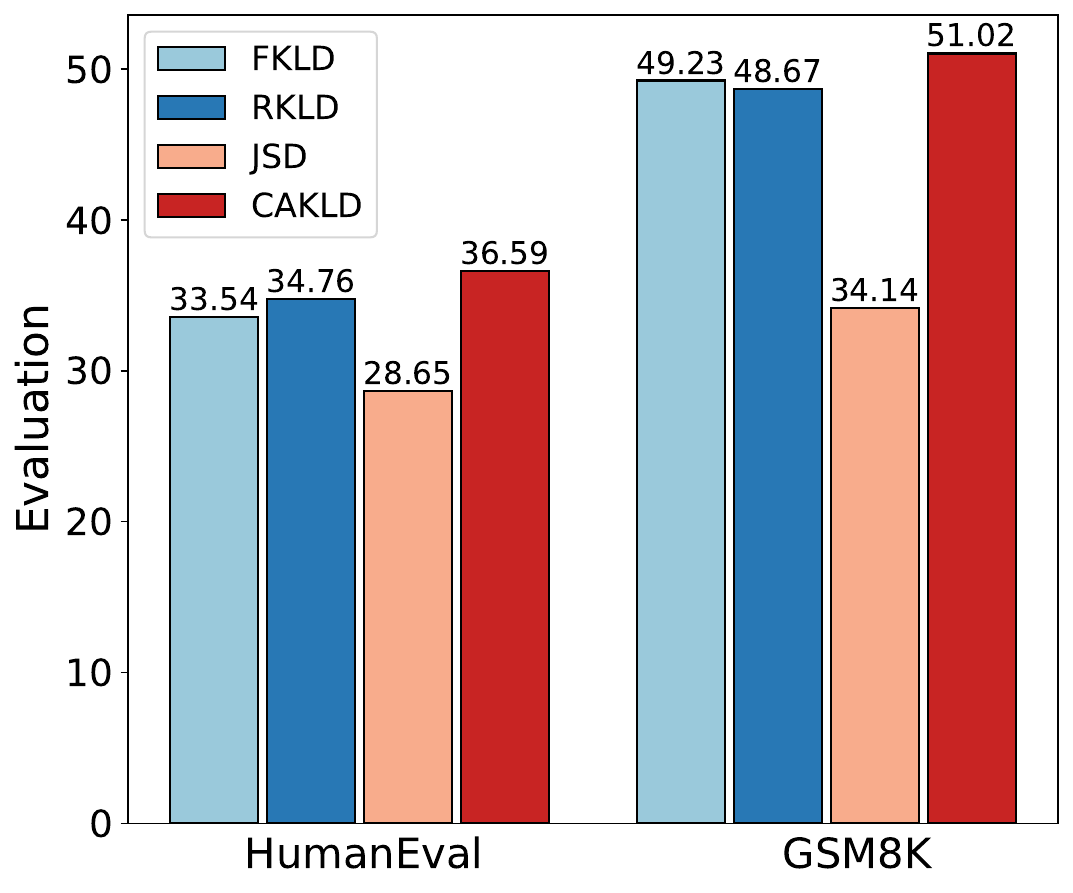} 
\caption{Performance comparison between different objective functions on WizardCoder-7B and MetaMath-7B with domain-specific tasks.}
\label{fig:objectives} 
\end{figure}

In Figure \ref{fig:objectives}, we demonstrate the effectiveness of our proposed Confidence-Aware KL Divergence (CAKLD) by showcasing performance indicators for reasoning tasks under different objective functions. Our findings show that CAKLD outperforms other objective functions. Though JSD also has a bounded coefficient for interpolation, in practice we observe that it has a weak ability to converge for QAT. 

\subsection{Analysis and Discussion}

\paragraph{Comparison with QuIP}

QuIP enhances 2-bit PTQ for LLMs through incoherence processing. Its subsequent iteration, QuIP\#\footnote{\url{https://cornell-relaxml.github.io/quip-sharp/}}, refines this approach by shifting from scalar quantization to vector quantization via lattice codebooks, significantly narrowing the performance gap with 16-bit models.
For a consistent comparison, we utilize the BF16 pretrained model and then apply Quip(\#) and BitDistiller.
As shown in Table~\ref{tab:compare_with_quip}, our BitDistiller surpasses QuIP across all benchmarks. In comparison with QuIP\#, BitDistiller retains its superior performance in language modeling and programming, while QuIP\# outperforms in mathematical reasoning. Being orthogonal to QAT with distillation, PTQ incorporating incoherence processing and vector quantization could potentially serve as an effective initialization method for BitDistiller. We intend to explore whether the integration of QuIP(\#) into BitDistiller can further improve the performance of low-bit models.
%\TODO{for shijie: explaine BF16?}

\begin{table}[h]
    \centering
    \resizebox{\columnwidth}{!}{
    \begin{tabular}{cc|ccc|c|c}
    \toprule
        \multicolumn{2}{c|}{\multirow{2}{*}{\textbf{Method}}} & \multicolumn{3}{c|}{\textbf{LLaMA-2-7B}}  & \textbf{WizardCoder-7B} & \textbf{MetaMath-7B} \\ 
        &  & PPL$\downarrow$ & MMLU (5s) & QA-avg & HumanEval & GSM8K \\ \midrule\midrule
        \multicolumn{2}{c|}{BF16} & 5.47 & 46.45 & 61.67 & 54.88 & 66.41 \\ \midrule
        \multirow{2}{*}{2 Bits} & Quip & 728.15 & 24.30 & 38.19 & 0.0 & 0.0 \\ 
        \multirow{2}{*}{g128} & Quip\# & 8.97 & 30.90 & 52.40 & 12.96 & 60.00 \\ 
        ~ & BitDistiller & 8.08 & 29.25 & 54.17 & 36.58  & 51.02 \\ 
        \bottomrule
    \end{tabular}
    }
    \caption{Performance comparison of 2-bit quantized models using QuIP, QuIP\#, and BitDistiller on LLaMA-2-7B, WizardCoder-7B, and MetaMath-7B.}
    \label{tab:compare_with_quip}
\end{table}

\paragraph{Comparison with TSLD}

%\shijie{do we need to explain why a distinct section about TSLD?}

\begin{figure}[h]  
    \centering  
    \begin{subfigure}{0.235\textwidth}  
        \includegraphics[width=\textwidth]{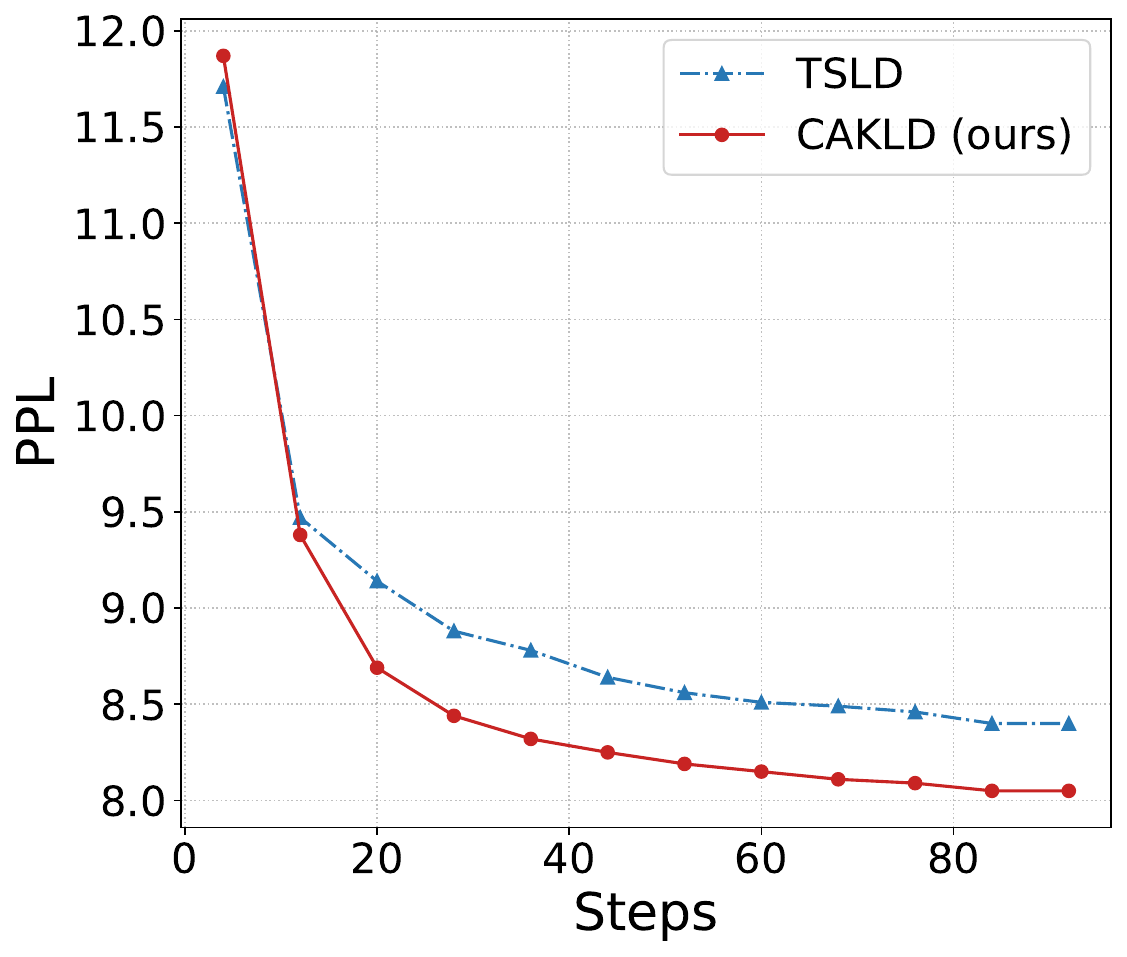}  
    \end{subfigure}  
    \begin{subfigure}{0.23\textwidth}  
        \includegraphics[width=\textwidth]{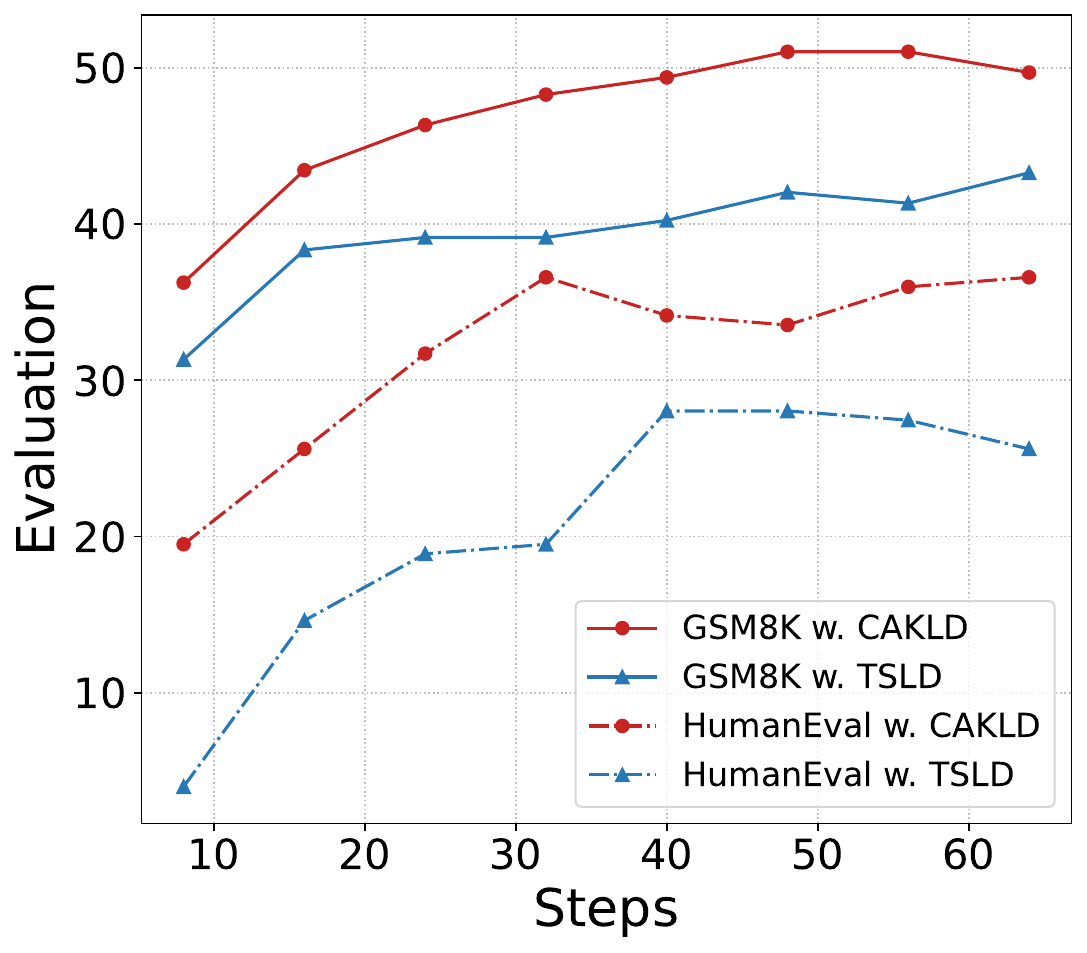}
    \end{subfigure}  
    \caption{Comparison of TSLD and CAKLD on perplexity (left) and reasoning tasks performance (right).} 
    \label{fig:compare_with_tsld} 
\end{figure}

Prior work~\cite{tsld} introduced Token-Scaled Logit Distillation (TSLD) to alleviate overfitting during QAT. 
To facilitate a direct and fair comparison between TSLD and our CAKLD, we incorporate TSLD into the BitDistiller framework by replacing CAKLD with TSLD while keeping all other settings unchanged.
As depicted in Figure \ref{fig:compare_with_tsld}, CAKLD not only converges more rapidly but also delivers superior overall performance compared to TSLD.

%The previous method, Token-Scaled Logit Distillation (TLSD) \cite{token-scale}, employs token-wise logit scaling to reduce overfitting in QAT. For a fair comparison, we adapted our implementation settings, altering only the loss function to contrast our CAKLD with Token-Scaled Logit Distillation. As observed in Figure \ref{fig:compare_with_tsld}, both loss functions significantly mitigate the occurrence of overfitting. However, CAKLD demonstrates a faster convergence and ultimately achieves better performance.

\paragraph{Effectiveness of Self-Distillation}
Table \ref{tab:self_distill} compares 2-bit QAT performance using the LLaMA-2-7B or larger LLaMA-2-13B as the teacher model. Surprisingly, in practice the larger 13B model didn't improve accuracy, hinting that a teacher with the same model architecture as the student may enhance weight alignment and probability distribution matching, thereby improving model effectiveness. 
Further investigation and deeper analysis are needed in future work to fully understand the implications of different teacher-student sizes and architectures in QAT.

\begin{table}[h]
    \centering
    \resizebox{\columnwidth}{!}{
    \begin{tabular}{c|c|c|c|c}
    \toprule
        \textbf{LLaMA-2-7B} & Quantized Student & Teacher & PPL $\downarrow$ & MMLU (5s) $\uparrow$ \\ \midrule \midrule
        
        2 Bits & 7B & 13B & 8.12 & 28.27 \\ 
        g128 & 7B & 7B & 8.08 & 29.25 \\ 
    \bottomrule
    \end{tabular}
    }
    \caption{Performance comparison of 2-bit quantized models using LLaMA-2-13B and LLaMA-2-7B as the teacher model.}
    \label{tab:self_distill}
\end{table}

\paragraph{Training Efficiency}

Table \ref{tab:training_efficiency} highlights the efficiency of BitDistiller compared to LLM-QAT \cite{llm-qat} in quantizing the WizardCoder-7B model. The results demonstrate a dramatic reduction in the total time required for quantization: BitDistiller completes the process in approximately 3 hours on a single A100-80G GPU, as opposed to the hundreds of GPU hours required by LLM-QAT. 
(Original LLM-QAT uses 64 GPUs. For a direct and fair comparison, we evaluate the GPU hours needed for LLM-QAT on a single GPU.)
%This stark contrast not only underscores the remarkable cost-efficiency of our method but also showcases its potential to expedite the quantization aware training process without sacrificing computational resources.

\begin{table}[h]
    \centering
    \resizebox{\columnwidth}{!}{
    \begin{tabular}{c|c|c|cccc}
    \toprule
        \multirow{2}{*}{\textbf{Method}} & \multirow{2}{*}{\textbf{Devices}} & \multirow{2}{*}{\textbf{\#Data}} & \multicolumn{4}{c}{\textbf{Time (Hours)}}  \\ 
        ~ & ~ & ~ & Data Gen & Quant Init & QAT & Total \\ \midrule \midrule
        LLM-QAT & \multirow{2}{*}{1 * A100 80G} & 100K & 270 & 0 & 10.64 & 280.64  \\ 
        BitDistiller &  & 2K & 1.47 & 0.63 & 0.92 & 3.02 \\ 
    \bottomrule
    \end{tabular}
    }
    \caption{Time required for LLM-QAT and BitDistiller to quantize WizardCoder-7B on a  NVIDIA A100-80G.}
    \label{tab:training_efficiency}
\end{table}

% \begin{table}[!ht]
%     \centering
%     \fontsize{8.5}{10}\selectfont
%     \begin{tabular}{c|c|c|c|c|c}
%     \hline
%         \multirow{2}{*}{Method} & \multirow{2}{*}{\#Data} & ~ & Time (Hours) & ~ & ~ \\ 
%         ~ & ~ & Data Generation & Quantization Initialization & QAT & Devices \\ \hline
%         LLM-QAT & 100K & 90 & 0 & ~ & 8 * A100 80G SXM \\ \hline
%         BitDistiller & 2K & ~ & ~ & ~ & 1 * A100 80G SXM \\ \hline
%     \end{tabular}
% \end{table}

\section{Conclusion}

BitDistiller leverages QAT with self-distillation to boost sub-4-bit LLM performance.
The asymmetric quantization and clipping strategies, coupled with the innovative CAKLD objective, facilitate faster learning and superior performance.
BitDistiller outperforms existing PTQ and QAT methods, achieving notable improvements in 3/2-bit settings across diverse language and reasoning tasks.
Moreover, BitDistiller is more cost-efficient with fewer data and training resources required.
%Looking ahead, we aim to adapt BitDistiller to the realm of 1-bit (binary) quantization, which is more challenging and also promising. 
\section*{Limitations}
%\TODO{extend to 1 bit and vector quant}

Despite the promising results demonstrated by BitDistiller, it is important to acknowledge certain limitations and areas for future investigation.

A key limitation lies in the empirical nature of our findings. For instance, the reason behind the counterintuitive outcome where a 7B model outperforms a 13B model as a teacher during the distillation of a 2-bit 7B student model. Having the same model architecture may be the reason but not detailed explained and understood. This highlights the need for a deeper investigation and theoretical exploration to complement our empirical observations.

Looking ahead, we aim to extend BitDistiller to the realm of 1-bit (binary) quantization. While this presents a more challenging scenario, it also offers the potential for significant advancements in efficient LLM inference as binary weights enables computation with only additions and without multiplications.

Moreover, the current iteration of BitDistiller applies exclusively to scalar quantization. As future work, we plan to explore the adaptation of BitDistiller to vector quantization. Preliminary research in this area indicates that vector quantization could yield substantial benefits, and incorporating it into our framework represents a natural and promising progression of our research.

%\section*{Ethics Statement}

\section*{Acknowledgements}
We would like to thank the HPC-AI-Integrated Intelligent
Computing center of HKUST(GZ) for providing some of the
hardware platforms in this project.

% Entries for the entire Anthology, followed by custom entries
\bibliography{anthology,reference}
\bibliographystyle{acl_natbib}

\appendix
\section{Appendix}
\label{sec:appendix}
\subsection{Details of PTQ and QAT Configuration}  
\label{subsec:ptq_qat_baselines}
\begin{figure}[ht] 
\centering 
\includegraphics[width=0.3\textwidth]{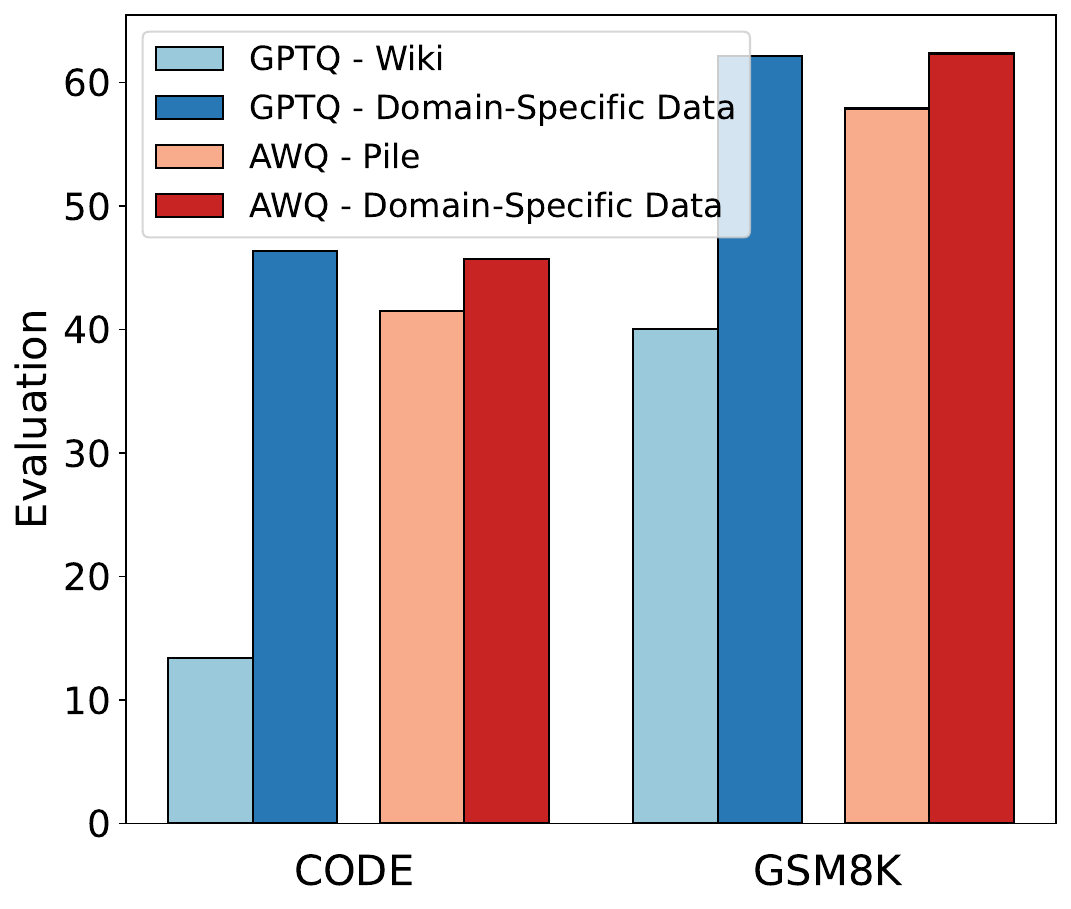} 
\caption{Comparative Evaluation of PTQ Methods Using Various Calibration Datasets}
\label{fig:calib_data} 
\end{figure}

We evaluate PTQ methods by examining the impact of different calibration dataset distributions. Illustrated in Figure \ref{fig:calib_data}, calibrating with domain-specific data significantly enhances task-specific performance. For a fair comparison, all PTQ methods utilize the default calibration datasets for general language tasks and domain-specific calibration datasets \cite{evol_teacher, metamath} for reasoning tasks. 

Regarding QAT methods, it should be noted that the use of symmetric quantization in LLM-QAT results in degradation when grouped quantization is applied. To ensure a fair comparison, we replicate the approach with our setup and employ asymmetric uniform quantization.

\subsection{Implementation Details and Analysis of Confidence-Aware KLD}
\label{subsec:cakld_detail}
\begin{figure}[h]  
    \centering  
    \begin{subfigure}{0.23\textwidth}  
        \includegraphics[width=\textwidth]{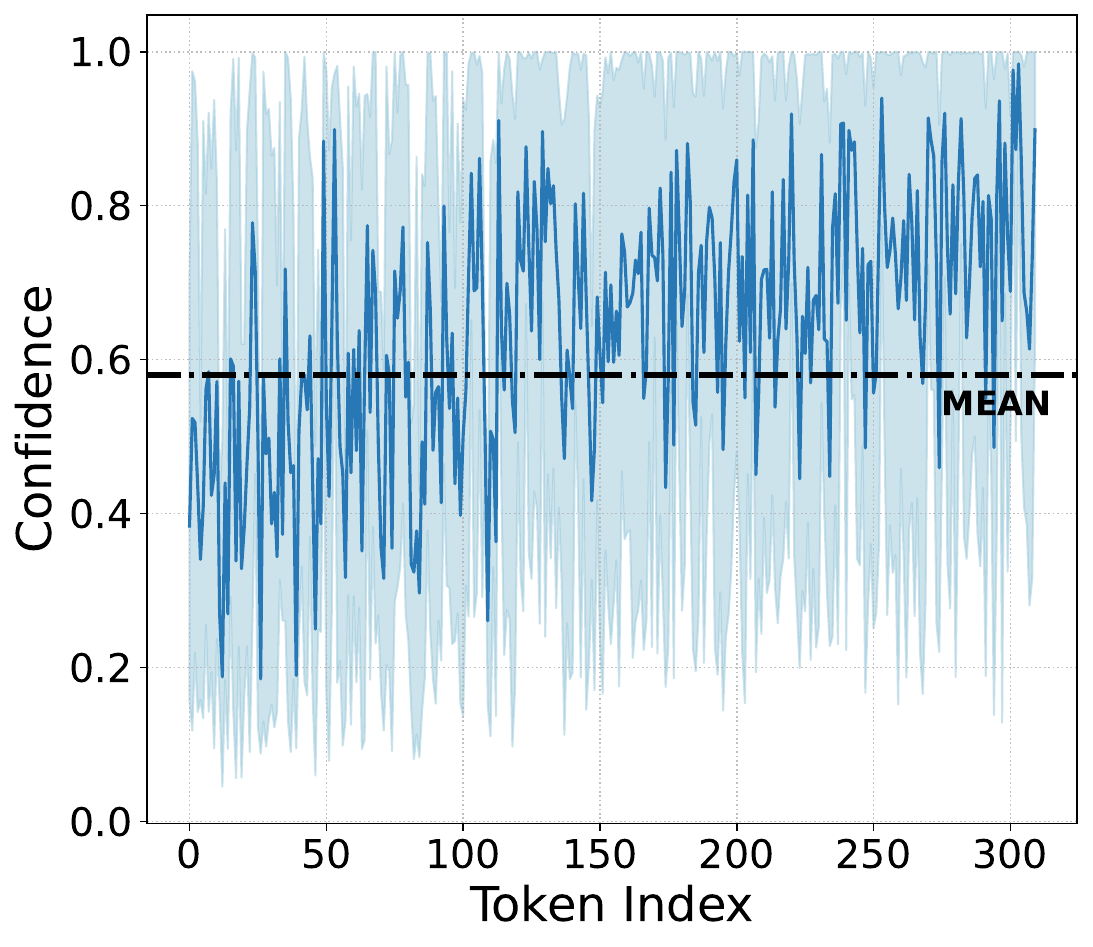} 
        \caption{Text Generation Task}
        \label{fig:conf_text_generation}
    \end{subfigure}  
    \begin{subfigure}{0.23\textwidth}  
        \includegraphics[width=\textwidth]{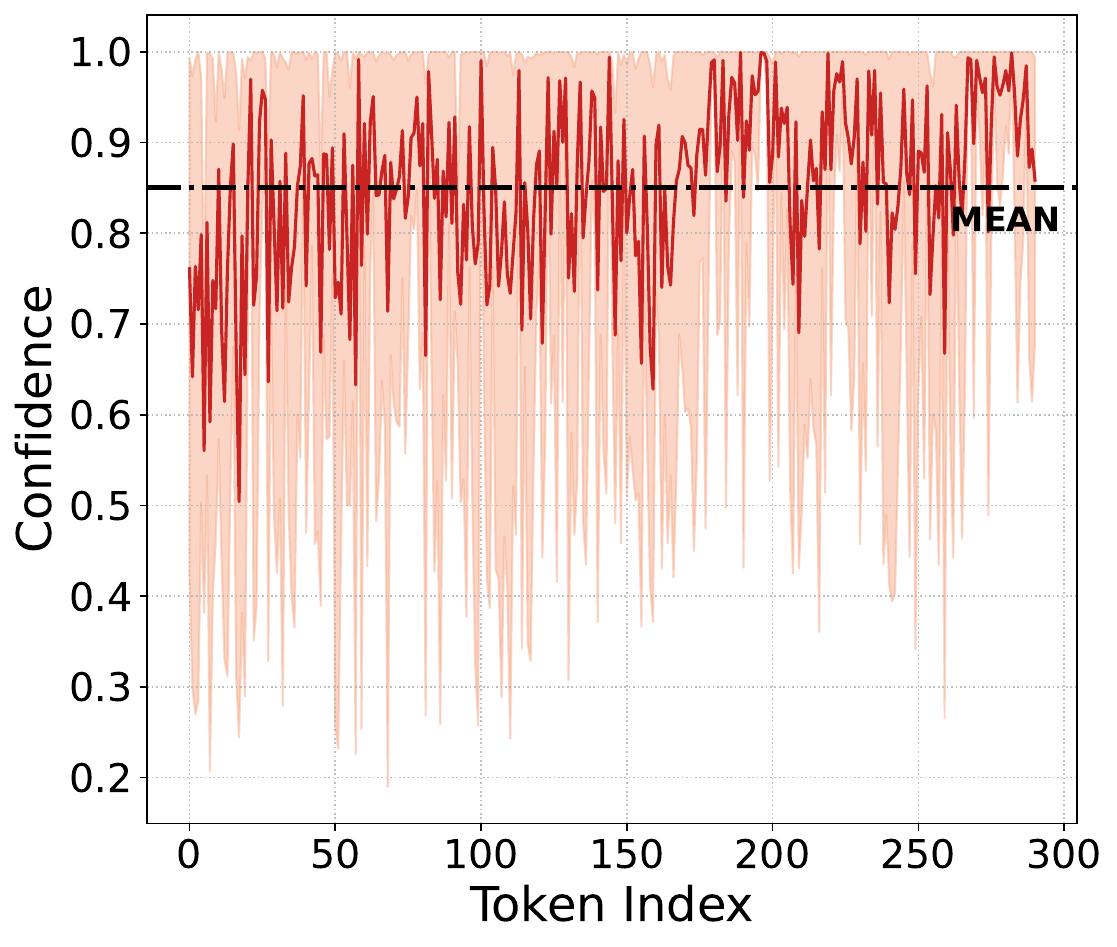}
        \caption{Reasoning Task}
        \label{fig:conf_reasoning}
    \end{subfigure}  
    \caption{Per-token confidence scores when teacher model (full-precision) conducting text generation task and reasoning task.} 
    \label{fig:confidence}
\end{figure} 
We use a straightforward method in the pre-calculation of the coefficient $\gamma$. We utilize ten batches of training data to perform forward passes without updating parameters. Subsequently, we obtain the logits from the teacher model to compute the average token probability. In Figure \ref{fig:confidence}, we have conducted analysis by examining the confidence scores of the teacher model in various tasks during next-word prediction. This analysis reveals that confidence levels can vary in text generation tasks, in contrast to reasoning tasks where each step is critical. Notably, in text generation tasks using LLMs, relying solely on the highest conditional probability through Greedy Search may result in local optima, overlooking more optimal sequences. These observations advocate for a mean-seeking Kullback-Leibler (KL) approach, encouraging the student model to encompass all potential modes of the teacher, thereby more effectively capturing the teacher’s general generative capabilities. In reasoning tasks, where the teacher model shows high confidence in next-word predictions, the student model should concentrate on learning the predominant mode from the teacher. Our proposed method, CAKLD, is designed to balance these two distinct modes effectively.

\subsection{Training Datasets Examples} \label{sec:training_data}
For general language tasks, we mix token sequences from Alpaca and WikiText-2 datasets with a ratio of 2:1. Since WikiText-2 lacks explicit instructions, we utilize the first 128 tokens from the corpus as the input prompt for the teacher model's generation process, setting the temperature to 0.7. For tasks related to code understanding and generation, we employ the Evol-Instruct-Code dataset. For mathematical reasoning, we utilize MetaMathQA. Examples of the training data utilized are shown in Table \ref{tab:training_data}. 

It is essential to highlight that our self-distillation process utilizes only a small portion of the involved datasets.

\subsection{Evaluation of General Language Tasks on LLaMA-2-13B} \label{subsec:appendix_13b}

Additional results of the General Language Tasks for LLaMA-2-13B are shown in Table \ref{tab:language_model_13b}.

\subsection{Integration with AWQ For Quantization Strategies}
\label{subsec:awq_init}
\begin{table}[h]
    \centering
    \resizebox{\columnwidth}{!}{
    \begin{tabular}{cc|c}
    \toprule
        \multicolumn{2}{c|}{\multirow{2}{*}{\textbf{LLaMA-2-7B}}} & PPL $\downarrow$ \\ 
        ~ & ~ & (start $\mapsto$ end ) \\ \midrule \midrule
        ~ & INT-Asym & 6.65 $\mapsto$ 6.15 \\ 
        3 Bits & AWQ & 6.48 $\mapsto$ 6.09 \\ 
        g128 & Clip-Asym & 6.21 $\mapsto$ 6.00 \\ 
        ~ & AWQ + Clip-Asym & 6.18 $\mapsto$ 6.00 \\ \midrule
        ~ & INT-Asym & 3.4e2 $\mapsto$ 16.94 \\ 
        2 Bits & AWQ & 2.2e5 $\mapsto$  Inf  \\ 
        g128 & Clip-Asym & 17.98 $\mapsto$ 8.08 \\ 
        ~ & AWQ + Clip-Asym & 16.61 $\mapsto$ 8.13 \\ 
    \bottomrule
    \end{tabular}
    }
    \caption{Results of quantization initialization for QAT combining with AWQ on PPL of WikiText-2.}
    \label{tab:quantization_awq}
\end{table}

As shown in Table \ref{tab:quantization_awq}, we explore the efficacy of combining asymmetric clipping with AWQ during the self-distillation process. Our results indicate that asymmetric clipping significantly enhances robustness in sub-4-bit quantization scenarios. For instance, at the 2-bit quantization level, both INT-Asym and AWQ methods are unable to complete the task. Conversely, Clip-Asym not only succeeds but also achieves a marked improvement in perplexity. It is also noteworthy that while integrating AWQ prior to QAT yields improvements initially, there is no additional performance gain after training. This suggests that a straightforward clipping approach is sufficiently effective for initializing QAT.

% \subsection{Post Training Quantization Comparison}
% \begin{table}[t]
%     \centering
%     \begin{tabular}{c|c|c|c|c}
%     \hline
%         LLaMA 2  & ~ & 7B & 13B & 70B \\ \hline
%         3 Bit & GPTQ & 6.38 & 5.41 & 3.92 \\ \hline
%         ~ & AWQ & 6.71 & 5.47 & 3.87 \\ \hline
%         ~ & Omniquant & 6.10 & 5.28 & 3.78 \\ \hline
%         ~ & Clip-Asym & 6.21 & 5.34 & 3.81 \\ \hline
%         2 Bit & GPTQ & NaN & 15.08 & 8.35 \\ \hline
%         ~ & AWQ & 2.2e5 & 1.2e5 & 7.3e4 \\ \hline
%         ~ & Omniquant & 12.84 & 9.15 & 7.82 \\ \hline
%         ~ & Clip-Asym & 17.98 & 9.74 & 5.88 \\ \hline
%     \end{tabular}
% \end{table}

\begin{table*}[t]
    \centering
    \scriptsize
    \begin{tabularx}{\textwidth}{c|c|c|>{\hsize=1.2\hsize}X|>{\hsize=0.8\hsize}X} % 确保 hsize 总和为 X 列数
    \toprule
        \textbf{Tasks} & \textbf{Resources} & \textbf{Number} & \textbf{Instruction} & \textbf{Teacher Response} \\ \midrule \midrule
        General Language Tasks & Alpaca & 4K & Find the five steps to meditate & 1. Find a quiet place to meditate$\backslash$n2. Sit in a comfortable position$\backslash$n3. Close your eyes and take a deep breath$\backslash$n4. Focus your attention on your breath$\backslash$n5. When thoughts invade your mind, gently bring your attention back to your breath \\ \midrule
        General Language Tasks & WikiText & 2K & In December 1856 , Rice brought forward two bills in Congress : an enabling act that would allow Minnesota to form a state cons & itution, and a bill that would allow Missouri to secede from the Union. The bills were defeated, and Rice was removed from the chairmanship of the House Judiciary Committee. ...... \\ \midrule \midrule
        Code-related Tasks & Evol-Instruct-Code & 2K & Create a nested loop to print every combination of numbers between 0-9, excluding any combination that contains the number 5. Additionally, exclude any combination that contains a repeating digit. Implement the solution without using any built-in functions or libraries to check for repeating digits. & Here's one way to implement the solution:$\textbackslash$n``` python$\textbackslash$r$\textbackslash$n for i in range(10):...... \\ \midrule \midrule
        Math-related Tasks & MetaMathQA & 2K & A store is comparing their profits throughout the year. They had profits of \$1,500 in the first quarter of the year, \$3,000 in the third quarter, and \$2,000 in the fourth quarter. If their annual profits are \$8,000, how much profit, in dollars, did they make in the second quarter? & $\backslash$nThe annual profit is the sum of the profits in each quarter, so \$8,000 = \$1,500 + \$3,000 + \$2,000 + x, where x is the profit in the second quarter.$\backslash$nTo find x, we need to isolate it on one side of the equation....... \\ 
        \bottomrule
    \end{tabularx}
    \caption{The Training Dataset examples for different tasks.}
    \label{tab:training_data}
\end{table*}

\begin{table*}[t]
    \centering
    \resizebox{\textwidth}{!}{
    %\begin{adjustbox}{margin=0.2em}
    \begin{tabular}{cc|c|c|cccc|c} 
    \toprule[\heavyrulewidth]
        \multicolumn{2}{c|}{\textbf{LLaMA-2-13B}} & PPL $\downarrow$ & MMLU (5s)  & PIQA & Hella. & Wino. & ARC-c & Avg \\ \midrule \midrule
        \multicolumn{2}{c|}{BF16} & 4.88 & 55.54 & 79.16 & 60.13 & 72.14 & 48.12 & 63.02 \\ \midrule
        ~ & RTN & 5.52 & 50.74 & 78.35 & 57.75 & 71.11 & 43.86 & 60.36 \\ 
        \multirow{2}{*}{3 Bits} & GPTQ & 5.41 & 50.63 & 77.26 & 56.84 & 70.72 & 42.83 & 59.66 \\ 
        \multirow{2}{*}{g128} & AWQ & 5.47 & 49.64 & 77.09 & 57.52 & 70.32 & 43.86 & 59.69 \\ 
        ~ & OmniQuant & 5.48 & 48.97 & 77.64 & 57.08 & 70.88 & 44.28 & 59.77 \\ 
        ~ & LLM-QAT & 5.32 & 51.60 & 78.29 & 58.45 & 70.56 & 44.62 & 60.70 \\ 
        ~ & \multicolumn{1}{>{\columncolor{mygray}}c|}{\textbf{BitDistiller (ours)}} & \multicolumn{1}{>{\columncolor{mygray}}c|}{\textbf{5.20}} & \multicolumn{1}{>{\columncolor{mygray}}c|}{\textbf{53.21}} & \multicolumn{1}{>{\columncolor{mygray}}c}{78.67} & \multicolumn{1}{>{\columncolor{mygray}}c}{58.66} & \multicolumn{1}{>{\columncolor{mygray}}c}{71.59} & \multicolumn{1}{>{\columncolor{mygray}}c|}{46.67} & \multicolumn{1}{>{\columncolor{mygray}}c}{\textbf{61.76}} \\ \midrule
        ~ & RTN & 109.21 & 24.74 & 57.56 & 32.56 & 50.75 & 21.84 & 37.49 \\
        ~ & GPTQ & 15.08 & 23.70 & 56.04 & 30.99 & 51.22 & 19.28 & 36.25 \\
        2 Bits & AWQ & 1.2e5 & 27.04 & 53.16 & 25.82 & 51.70 & 23.04 & 36.15 \\
        g128 & OmniQuant & 25.69 & 26.09 & 61.81 & 31.92 & 51.38 & 22.27 & 38.69 \\
        ~ & LLM-QAT & 7.80 & 29.37 & 74.10 & 49.49 & 63.14 & 33.87 & 49.99 \\ 
        ~ & \multicolumn{1}{>{\columncolor{mygray}}c|}{\textbf{BitDistiller (ours)}} & \multicolumn{1}{>{\columncolor{mygray}}c|}{\textbf{6.78}} & \multicolumn{1}{>{\columncolor{mygray}}c|}{\textbf{37.50}} & \multicolumn{1}{>{\columncolor{mygray}}c}{75.84} & \multicolumn{1}{>{\columncolor{mygray}}c}{51.30} & \multicolumn{1}{>{\columncolor{mygray}}c}{65.90} & \multicolumn{1}{>{\columncolor{mygray}}c|}{37.46} & \multicolumn{1}{>{\columncolor{mygray}}c}{\textbf{53.60}} \\
    \bottomrule
    \end{tabular}
    }
    %\end{adjustbox}
    \caption{\textbf{General language task} results of BitDistiller versus established PTQ and QAT Methods on LLaMA-2-13B Model. Our method achieves leading performance in both 3-bit and 2-bit quantization.}
    \label{tab:language_model_13b}
\end{table*}

% This is a section in the appendix.

% group size (+64
% 13B llama
% with GPTQ init, AWQ init
% llm-qat(original) llm-qat(asym)

% Figure3 as analysis

\end{document}